\documentclass[10pt,twocolumn,letterpaper]{article}

\usepackage{iccv}
\usepackage{times}
\usepackage{epsfig}
\usepackage{graphicx}
\usepackage{amsmath}
\usepackage{amssymb}

\usepackage{multirow}
\usepackage{multicol}
\usepackage{color}
\usepackage{bbding}
\usepackage{colortbl}
\usepackage{soul}
\definecolor{Gray}{gray}{0.9}
\sethlcolor{Gray}
\usepackage{subcaption}
\usepackage{diagbox}

\usepackage[pagebackref=true,breaklinks=true,letterpaper=true,colorlinks,bookmarks=false]{hyperref}

\iccvfinalcopy % *** Uncomment this line for the final submission

\ificcvfinal\fi

\begin{document}

%%%%%%%%% TITLE
\title{SMMix: Self-Motivated Image Mixing for Vision Transformers}

\author{Mengzhao Chen$^{1}$, Mingbao Lin$^3$, Zhihang Lin$^1$, Yuxin Zhang$^1$, Fei Chao$^{1}$, Rongrong Ji$^{1,2}$\thanks{Corresponding Author}\\
$^1$MAC Lab, School of Informatics, Xiamen University \\
$^2$Institute of Artificial Intelligence, Xiamen University\\
$^3$Tencent Youtu Lab\\
{\tt\small \{cmzxmu, lmbxmu, yuxinzhang, zhihanglin\}@stu.xmu.edu.cn, \{fchao, rrji\}@xmu.edu.cn}
}

\maketitle
% Remove page # from the first page of camera-ready.
\ificcvfinal\thispagestyle{empty}\fi

\begin{abstract}
CutMix is a vital augmentation strategy that determines the performance and generalization ability of vision transformers (ViTs). However, the inconsistency between the mixed images and the corresponding labels harms its efficacy. Existing CutMix variants tackle this problem by generating more consistent mixed images or more precise mixed labels, but inevitably introduce heavy training overhead or require extra information, undermining ease of use.
To this end, we propose an novel and effective Self-Motivated image Mixing method (SMMix), which motivates both image and label enhancement by the model under training itself. Specifically, we propose a max-min attention region mixing approach that enriches the attention-focused objects in the mixed images. Then, we introduce a fine-grained label assignment technique 
that co-trains the output tokens of mixed images with fine-grained supervision.
Moreover, we devise a novel feature consistency constraint to align features from mixed and unmixed images.
Due to the subtle designs of the self-motivated paradigm, our SMMix is significant in its smaller training overhead and better performance than other CutMix variants. In particular, SMMix improves the accuracy of DeiT-T/S/B, CaiT-XXS-24/36, and PVT-T/S/M/L by more than +1\% on ImageNet-1k. The generalization capability of our method is also demonstrated on downstream tasks and out-of-distribution datasets. Our project is anonymously available at \url{https://github.com/ChenMnZ/SMMix}.
   
\end{abstract}

%%%%%%%%% BODY TEXT
\section{Introduction}

\begin{figure}[!t]
    \centering
    \includegraphics[width=0.8\linewidth]{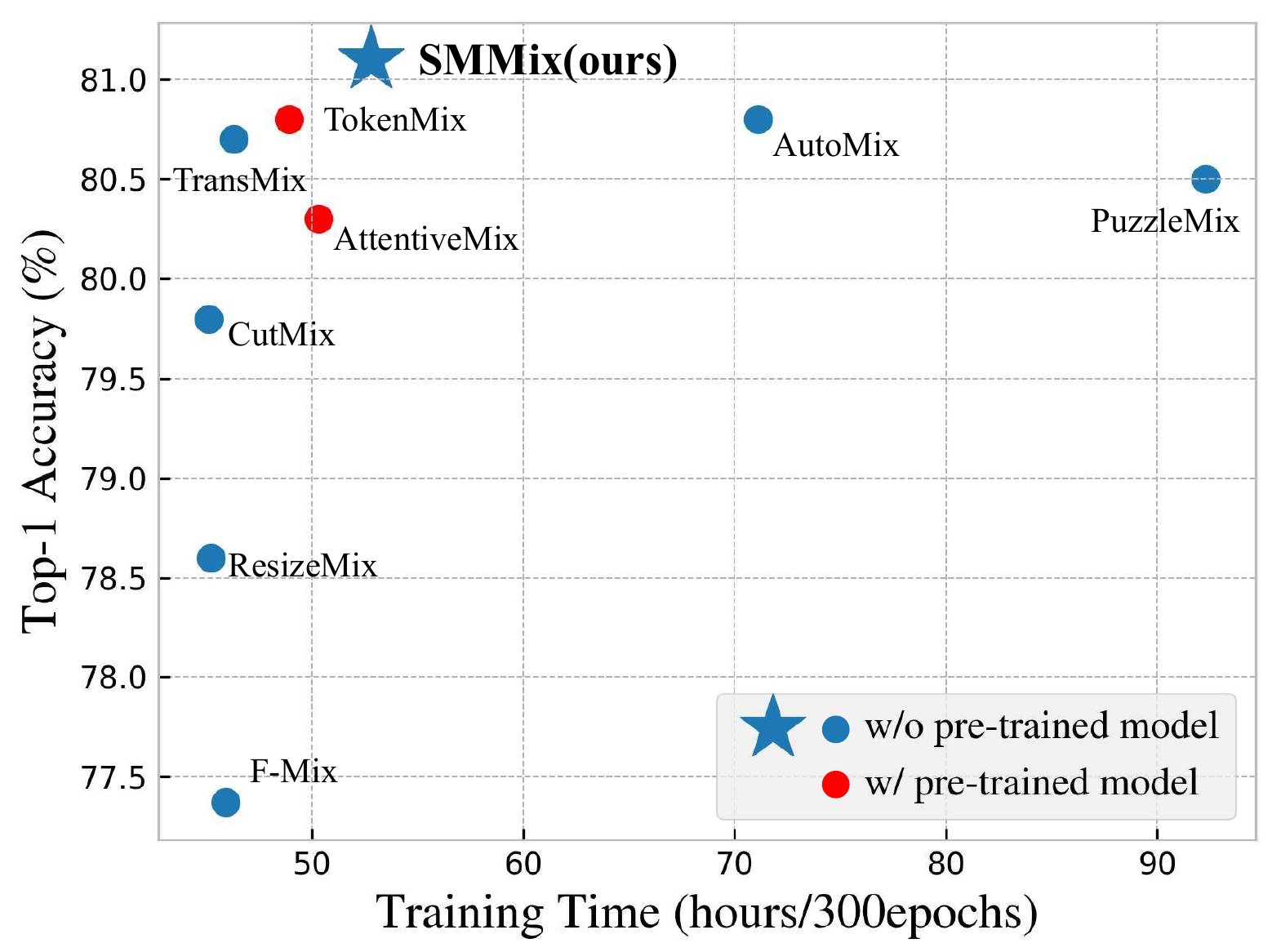}
    % \vspace{-1em}
    \caption{Training time \emph{vs.} accuracy with DeiT-S on ImageNet-1k. SMMix outperforms existing methods with light overhead.}
    \label{fig:efficiency}
    \vspace{-1em}
\end{figure}

Vision transformers (ViTs)~\cite{dosovitskiy2020image} have made substantial breakthroughs across various vision tasks, such as classification~\cite{dosovitskiy2020image,touvron2021training,wang2021not,jiang2021all,si2022inception,chen2021crossvit}, detection~\cite{carion2020end,zhu2020deformable,li2022dn,gao2022adamixer}, and segmentation~\cite{zheng2021rethinking,strudel2021segmenter,xie2021segformer,gu2022multi}.
However, the data-hungry problem~\cite{dosovitskiy2020image,touvron2021training} of ViT causes a serious overfitting problem when the data is insufficient.
%TODO
In order to improve the generalization of ViTs, data mixing augmentation techniques such as Mixup~\cite{zhang2017mixup} and CutMix~\cite{yun2019cutmix}, are used in the ViTs training recipe.
%Data augmentation, one of the most effective problem-solvers, has been extensively excavated. 
In particular, CutMix randomly crops a patch from the source image, pastes it into the target image, and forms a ground-truth label by mixing the labels of the source and target images in proportion to the area ratio of the mixed image. CutMix~\cite{yun2019cutmix} has been demonstrated to greatly enhance the generalization of ViTs. For example, CutMix increases the top-1 accuracy of ViT-Small~\cite{dosovitskiy2020image} by 4.1\%~\cite{li2022openmixup} on ImageNet-1k~\cite{deng2009imagenet} validation set. 
%when classifying the ImageNet-1k dataset~\cite{deng2009imagenet}.

%

\begin{figure}[!t]
\centering
\includegraphics[width=0.95\linewidth,height=0.7\linewidth]{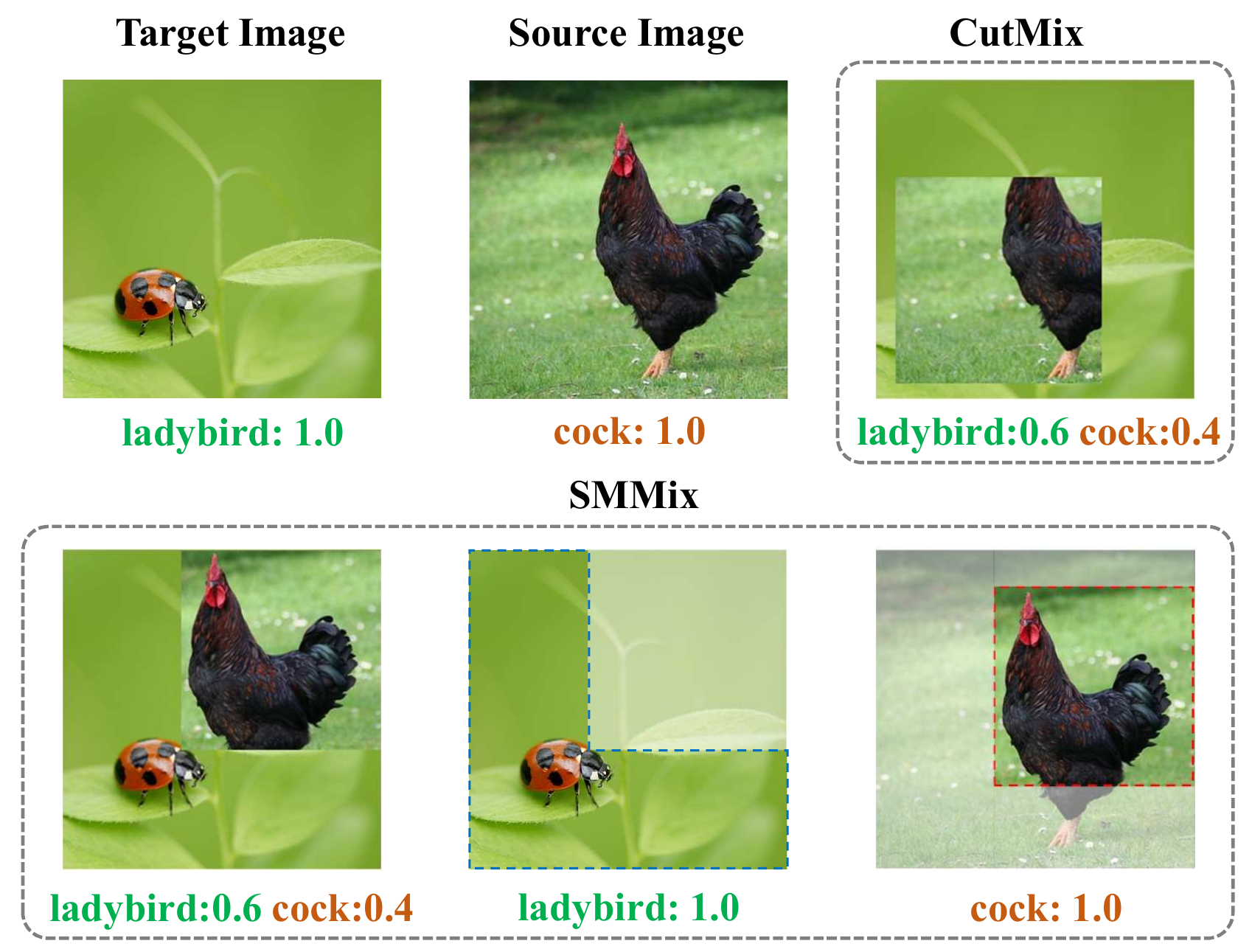}
\vspace{-0.5em}
\caption{CutMix \emph{vs.} SMMix. CutMix pastes a randomly-cropped patch in the source image to the target image, while SMMix pastes the attentive region in the source image to the unattentive  region in the target image. SMMix is co-trained by three fine-grained labels instead of a simple mixed label as CutMix.}
\label{fig:mix_visual} 
\vspace{-1.5em}
\end{figure}

Despite the progress, the image-label inconsistency issue also stems from the random patch selection and linear label combination.
Figure\,\ref{fig:mix_visual} illustrates a typical example, in which the mixed image of CutMix does not contain any hints of ladybirds. However, ladybird still appears on the generated mixed label. 
Such an image-label inconsistency issue prevents ViTs  from further improving performance.
Two mainstream methods: 1) image-driven~\cite{uddin2020saliencymix,kim2020puzzle,walawalkar2020attentive,liu2021unveiling} and 2) label-driven~\cite{chen2022transmix,liu2022tokenmix,jiang2021all,yun2021re,liu2022decoupled}, have recently been considered to overcome the drawbacks of CutMix.
The former method is dedicated to enhancing the saliency of mixed images, while the latter method aims to enhance the precision of mixed labels.
Nevertheless, these methods usually come with heavy training overhead, such as requiring pre-trained models~\cite{walawalkar2020attentive,liu2022tokenmix,jiang2021all,yun2021re}, double forward and backward propagations~\cite{kim2021co,kim2020puzzle,liu2021unveiling}, or additional generators~\cite{liu2021unveiling}, which may undermine the ease of the use of image mixing technique.
Moreover, these methods only consider the image and label enhancement in isolation, resulting in limited efficiency.

To address the aforementioned challenges, we propose a novel method, \textbf{S}elf-\textbf{M}otivated image \textbf{Mix}ing (SMMix), to enhance image mixing with ViTs. 
By leveraging the bootstrapping capabilities of the model under training itself, SMMix simultaneously motivates image and label enhancement with light training overhead.
Specially, we first use the image attention score in Eq.\,(\ref{eq:image_attention_score}) that accumulates attention score across all the image tokens. 
The motivations are from a widely-accepted actuality in existing works~\cite{liang2022not,xu2022evovit,chen2022coarse}, in which the class attention score from the self-attention operation can locate semantic objects. 
Therefore, we opt to use the image attention score to extend the general applicability of SMMix, since class attention is often unavailable for ViT models without a class token, while the image attention score can be easily obtained by feeding original images to a ViT model. 
With the guidance of the image attention score, we select the maximum-scored (most attentive) region from a source image and paste it to the region with a minimum attention score in a target image. We term this process as \textit{max-min attention region mixing}, which alleviates the image-label inconsistency issue by 
enriching the attention-focused objects in mixed images.

Distinctive from the prerequisite to tuning mixed labels~\cite{chen2022transmix,liu2022tokenmix}, capturing attentive objects in mixed images allows for a \textit{fine-grained label assignment}. 
We supervise different regions in a mixed image with different labels.
Concretely, the output tokens of a mixed image are assigned three types of labels to accomplish the label enhancement, as illustrated in Figure\,\ref{fig:mix_visual}, including mixed image label, target image label, and source image label.
We aggregate all output tokens, the result is then supervised by the mixed image label.
We also use region-specific supervision, \emph{i.e.}, target image labels and source image labels, to supervise the aggregated results of tokens from the target regions and source regions, respectively.
%
%In this way, we improve the recognition capabilities of ViTs with negligible computational costs.
%Note that, TokenLabel~\cite{jiang2021all} considers a similar fine-grained label by assigning each token a location-specific supervision, which however, heavily relies upon an over-parameterized NAFNet-F6 model~\cite{brock2021high}.%(438M parameters)~\cite{brock2021high}.

With label-consistent mixed images, we can extract mixed image features from ViTs. To correctly recognize the mixed images, we expect the features of mixed images to fall into a consistent space with those of original unmixed images.
We realize this function by creating a \textit{feature consistency constraint}, which aligns the feature distributions between mixed images and the linear combination of unmixed images.
Specially, SMMix can obtain the feature distributions and the image attention score of unmixed images from the same forward propagation of the model under training, resulting in light overhead.

Based on the above considerations, three key components are proposed in this paper, including 1) max-min attention region mixing (Sec.\,\ref{sec:max-min_attention_region_mixing}), 2) fine-grained label assignment (Sec.\,\ref{sec:fine_grained_label_assignment}), and 3) feature consistency constraint (Sec.\,\ref{sec:feature_consistency_constraint}). We term our method \textit{self-motivated image mixing}, since these components eliminate the dependency on pre-trained models and simply depend on the model under training itself.
%
%Table\,\ref{tab:module_ablation} shows the efficacy of each component.
%
%{\color{blue} Table\,\ref{tab:region_level_attention_score} shows that ViT-based models pre-trained by the proposed SMMix can generate more appropriate image attention score.}
%
We have performed extensive experiments, which demonstrate the powerful ability of our SMMix to boost the performance of various ViT-based models, including DeiT~\cite{touvron2021training} with a plain architecture, PVT~\cite{wang2021pyramid} with a hierarchical architecture, CaiT~\cite{touvron2021going} with deeper depth, and Swin~\cite{liu2021swin} with local self-attention. 
Moreover, our SMMix achieves better training overhead and performance trade-off because of the self-motivated paradigm. As shown in Figure\,\ref{fig:efficiency},
our SMMix can achieve state-of-the-art top-1 accuracy with light training overhead and does not require pre-trained models.

\section{Related Work}

\subsection{Vision Transformers}
%
%Vision Transformers (ViTs)~\cite{dosovitskiy2020image} conduct a patch-splitting operation and linear projection to flatten images to form a token sequence which is then fed into the transformer blocks. Each transformer block consists of a multi-head self-attention module (MHSA) and a feedforward network (FFN). 
Vision Transformer (ViT)~\cite{dosovitskiy2020image} shows the visual recognition ability of an original transformer~\cite{vaswani2017attention}.
% DeiT
However, ViT is easier to overfit on small datasets due to the lack of inductive bias. To handle this problem, DeiT~\cite{touvron2021training} introduces a powerful training recipe with various data augmentations~\cite{yun2019cutmix,zhang2017mixup,cubuk2020randaugment} and regularization techniques~\cite{huang2016deep,hoffer2020augment,zhong2020random}. 
% Variant
%
Based on the DeiT~\cite{touvron2021training} training recipe, many ViT-based architectures~\cite{liu2021swin,si2022inception,chu2021twins,dong2022cswin,ren2022shunted,heo2021rethinking,chen2021visformer,chen2021crossvit,touvron2021going,wang2021pyramid} are proposed to improve performance on various vision tasks.
In this work, we focus on improving CutMix~\cite{yun2019cutmix} augmentation, one of the data augmentation methods in DeiT~\cite{touvron2021training} training recipe. 

\subsection{Variants of CutMix}
CutMix~\cite{yun2019cutmix} randomly crops a patch from the source image and pastes it to the same location in the target image. Ground-truth labels of mixed images are generated by linearly combining the labels of the source and target images in proportion to the area ratio of the mixed images. 
However, such random crop-and-paste may cause image-label inconsistency. Existing works try to alleviate such a problem from two perspectives as follows.

\begin{figure*}[!t]
\centering
\includegraphics[width=0.9\textwidth]{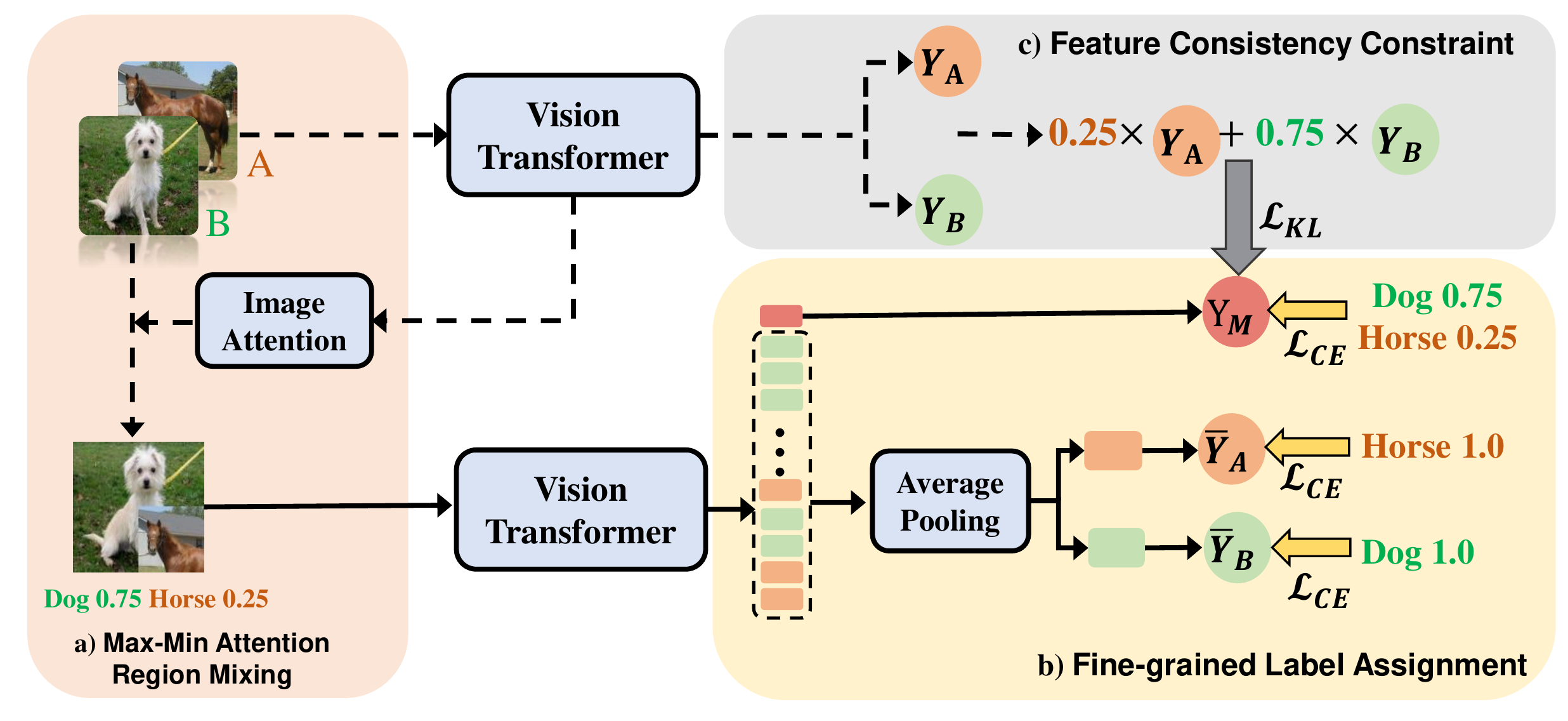}
\vspace{-0.5em}
\caption{Framework of SMMix that contains three components: a) max-min attention region mixing: maximizing the information of mixed images according to the image attention score. b) fine-grained label assignment: applying different supervision labels to tokens from different regions. c) feature consistency constraint: restraining the model to extract consistent features for mixed and unmixed images. 
%We just forward the model with unmixed images to get the corresponding image attention score and prediction distributions. Then train the model with the three proposed components. Besides,
The dashed arrows indicate no backward propagation. (Best viewed in colors)}
\label{fig:framework} 
\vspace{-1.0em}
\end{figure*}

% \paragraph{Image-driven Reconstruction}
\textbf{Image-driven Reconstruction.}
Image-driven reconstruction methods aim to maximize the saliency of mixed images. SaliencyMix~\cite{uddin2020saliencymix} and AttetiveMix~\cite{walawalkar2020attentive} select the cropped patches based on the saliency maps, which are obtained by statistical saliency model or pre-trained model. Furthermore, based on double forward and backward propagations, PuzzleMix~\cite{kim2020puzzle} and Co-Mixup~\cite{kim2021co} find the optimal mixed mask by solving the combinatorial optimization problem. Recently, instead of manually designing the mixing policies, AutoMix~\cite{liu2021unveiling} trains an additional mixup generator to generate mixed samples. 
As can be seen, the strategies for maximizing the saliency of mixed images are becoming increasingly sophisticated. To address such problems, our SMMix proposes a simple yet effective max-min attention region mixing to enhance the mixed images.

\textbf{Label-driven Reconstruction.}
Label-driven reconstruction methods dedicate to
generating more precise labels.
TransMix~\cite{chen2022transmix} mixes labels based on the class attention score. Other works~\cite{liu2022tokenmix,jiang2021all,yun2021re} rely on a big-scale model pre-trained on JFT-300M~\cite{sun2017revisiting}.
Based on the activation map of pre-trained model, TokenMix~\cite{liu2022tokenmix} assigns content-based mixes labels to mixed images, TokenLabel~\cite{jiang2021all} generates token-level supervision, and ReLabel~\cite{yun2021re} reorganizes the ImageNet-1k training set into a multi-label framework. 
Instead of depending on pre-trained models and adjusting the mixed labels, our SMMix proposes fine-grained label assignment, which provides fine-grained supervision to the output tokens by ground-truth labels.

\section{Preliminary}

\subsection{CutMix Augmentation}
CutMix~\cite{yun2019cutmix} enhances data diversity by mixing images. Let $\mathbf{x}$ and $\mathbf{y}$ denote a training image and its label, where $\mathbf{x} \in \mathbb{R}^{H \times W \times C}$. Given a source image-label training pair $(\mathbf{x}_A,\mathbf{y}_A)$ and a target one $(\mathbf{x}_B,\mathbf{y}_B)$, CutMix generates a new training sample $(\mathbf{\Tilde{x}},\mathbf{\Tilde{y}})$ as follows:
\begin{align}\label{cutmix_mix}
    \Tilde{\mathbf{x}} &= \mathbf{M} \odot \mathbf{x}_A + (\mathbf{1}-\mathbf{M}) \odot \mathbf{x}_B, \nonumber \\
    \Tilde{\mathbf{y}} &= \lambda \mathbf{y}_A + (1-\lambda) \mathbf{y}_B,
\end{align}
where $\mathbf{M} \in \{0,1\}^{H \times W}$ denotes a rectangular binary mask that indicates where to drop or keep in the two images, $\odot$ is element-wise multiplication, and $\lambda$ is the combination ratio sampled from a beta distribution. Note that $\lambda$ indicates the area ratio of $\mathbf{x}_A$ in mixed image $\mathbf{\Tilde{x}}$, \emph{i.e.}, $\lambda = \frac{\sum \mathbf{M}}{HW}$.

\subsection{Vision Transformer}
\textbf{Loss Computing.}
Given a ViT-based model $\mathcal{V}$, the output token sequence of an input image $\mathbf{x}$ is:
\begin{equation}
    \mathcal{V}(\mathbf{x}) = [(\mathbf{X}^{cls}); \mathbf{X}^1;...;\mathbf{X}^N],
\end{equation}
where $N$ is the total number of image tokens, $X^i$ is the $i$-th image token, and $X^{cls}$ corresponds to the class token, which exists only in some of the ViT-based architectures~\cite{touvron2021training,wang2021pyramid, touvron2021going}. The final prediction distribution $\mathbf{Y}$ is obtained with a classifier $\mathcal{F}$:
% \begin{align}
%  \mathbf{Y} = \mathcal{F}(X_cls), \text{w/ class token} ; \nonumber \\
%  \mathbf{Y} = \mathcal{F}(\frac{1}{N}\sum_{i=1}^{N}X^i),  \text{w/o class token} ,
% \end{align}
\begin{equation}
\mathbf{Y}=\left\{
         \begin{array}{lrr}
         \mathcal{F}(\mathbf{X}^{cls})&, \text{w/ class token},&   \\
         \mathcal{F}(\frac{1}{N}\sum_{i=1}^{N}\mathbf{X}^i)&,\text{w/o class token}.&  
         \end{array}
\right.
\end{equation}

The classification loss for the image $\mathbf{x}$ is:
\begin{equation}
    L = CE(\mathbf{Y},\mathbf{y}),
\end{equation}
where $CE(\cdot,\cdot)$ represents the cross entropy function.

\textbf{Self-Attention Operation.}
Self-attention operation is the key component of ViT. Given an image token sequence\footnote{We take the case without class token as an example here.} $\mathbf{T} \in \mathbb{R}^{N \times d}$. It is firstly linearly mapped into three matrices, namely $\mathbf{Q}$, $\mathbf{K}$ and $\mathbf{V}$. Then, the self-attention operation is computed as:
\begin{gather}
    \mathcal{A}(\mathbf{Q},\mathbf{K}) = \mathrm{Softmax}(\frac{\mathbf{Q} \mathbf{K}^{T}}{\sqrt{d}}) = [\mathbf{A}^1;\mathbf{A}^2;...;\mathbf{A}^N], \nonumber \\
    \mathrm{Attention}(\mathbf{Q},\mathbf{K},\mathbf{V}) = \mathcal{A}(\mathbf{Q},\mathbf{K})\mathbf{V}.
\end{gather}

The image attention score $\boldsymbol{\alpha} \in \mathbb{R}^N$ is derived as:
\begin{equation}\label{eq:image_attention_score}
    \boldsymbol{\alpha} = \frac{1}{N}\sum_{i=1}^{N}\mathbf{A}^i = [{\alpha}^1,{\alpha}^2,...,{\alpha}^N].
\end{equation}

The image attention score above is the result of single-head self-attention. For multi-head self-attention, we simply average across all attention heads to get the final image attention score.

\section{Self-Motivated Image Mixing}
This section formally introduces our SMMix, a novel image mixing method that maximizes the information of the mixed image and provides more fine-grained labels. 
Figure\,\ref{fig:framework} illustrates an overview of our proposed SMMix. Details are given below.

\subsection{Max-Min Attention Region Mixing}\label{sec:max-min_attention_region_mixing}

\begin{figure}[!t]
    \centering
    \includegraphics[width=0.95\linewidth]{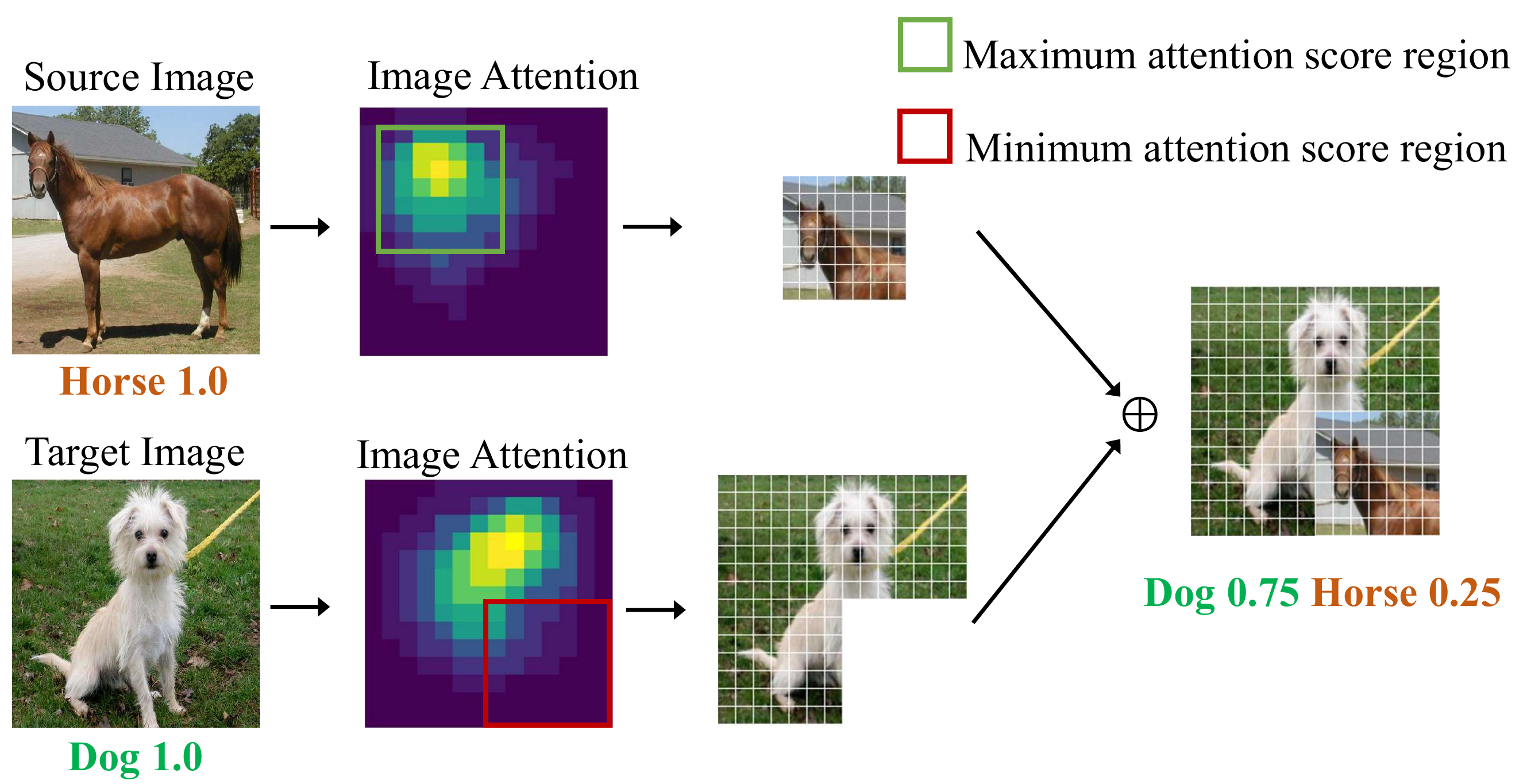}
    \caption{The pipeline of max-min attention region mixing.}
    \label{fig:max_min_mix_example}
    \vspace{-1.5em}
\end{figure}

To maximize the information of mixed images, our max-min attention region mixing replaces a minimum-scored region of the target image with a maximum-scored region of the source image.
As depicted in Figure\,\ref{fig:max_min_mix_example}, we split the source image $\mathbf{x}_A$ and the target image $\mathbf{x}_B$ into non-overlapping patches of size $P \times P$. A total of $N=\frac{H}{P} \times \frac{W}{P}$ patches are obtained for each image. Therefore, $\mathbf{x}_{A}$ and $\mathbf{x}_{B}$ are reorganized as $\mathbf{x}_{A},\mathbf{x}_{B} \in \mathbb{R}^{\frac{H}{P} \times \frac{W}{P} \times (P^2C)}$, row of which corresponds to a token. 
Then, we feed them into a ViT model to get the corresponding image attention scores $\boldsymbol{\alpha}_A \in \mathbb{R}^N$ and $\boldsymbol{\alpha}_B \in \mathbb{R}^N$.
Similarly, we rearrange the shape of their image attention score vectors, $\boldsymbol{\alpha}_{A}$ and $\boldsymbol{\alpha}_{B}$, to matrices of $\frac{H}{P} \times \frac{W}{P}$.

Similarly to CutMix, we intend to crop a region from the source image and paste the region into the target image to form a mixed image. To this effect, we introduce a side ratio $\delta$, sampled from a uniform distribution (0.25, 0.75), to determine the total $\lfloor \delta \frac{H}{P} \rfloor \times \lfloor \delta  \frac{W}{P} \rfloor$ patches within the cropped region. Our core difference in this paper is to locate the most informative region in the source image, and the least informative region in the target image. 
Concretely, the center indices of these two regions are defined as:
\begin{gather}
    i_s,j_s = \mathop{\arg\max}\limits_{i,j} \sum_{p,q} \boldsymbol{\alpha}_{A}^{i+p-\lfloor \frac{h}{2} \rfloor, j+q-\lfloor \frac{w}{2} \rfloor}, \nonumber \\
    i_t,j_t = \mathop{\arg\min}\limits_{i,j} \sum_{p,q} \boldsymbol{\alpha}_{B}^{i+p-\lfloor \frac{h}{2} \rfloor, j+q-\lfloor \frac{w}{2} \rfloor}, \label{eq:center_index}
\end{gather}
where $h=\lfloor \delta \frac{H}{P} \rfloor$, $w=\lfloor \delta  \frac{W}{P} \rfloor$, $p \in \{0,1,...,h-1 \}$, and $q \in \{0,1,...,w-1 \}$. 
It is intuitive that the selected region contains patches with the maximum attention score of the source image and the minimum attention score of the target image.

Then, in contrast to CutMix of Eq.\,(\ref{cutmix_mix}), we obtain the new mixed training sample $(\mathbf{{x}}_{M},\mathbf{{y}}_{M})$ as follows:
\begin{gather}
    {\mathbf{x}}_{M} = \mathbf{x}_{B},  \nonumber \\
    {\mathbf{x}}_{M}^{i_t+p-\lfloor \frac{h}{2} \rfloor, j_t+q-\lfloor \frac{w}{2} \rfloor} = \mathbf{x}_{A}^{i_s+p-\lfloor \frac{h}{2} \rfloor, j_s+q-\lfloor \frac{w}{2} \rfloor}, \nonumber \\
    \mathbf{{y}}_{M} = \lambda_M\mathbf{y}_A + (1-\lambda_M)\mathbf{y}_B,
\end{gather}
where $\lambda_M=\frac{hwP^2}{HW}$. 

\subsection{Fine-grained Label Assignment}\label{sec:fine_grained_label_assignment}
We feed the mixed image $\mathbf{x}_M$ to the ViT model to obtain the final output image token sequence $\mathbf{X}_M = [\mathbf{X}^{1}; \mathbf{X}^2;...;\mathbf{X}^N]$ and prediction distribution $\mathbf{Y}_M$.  Then, the traditional classification loss is:
\begin{equation}\label{classification}
    L_{cls} = CE(\mathbf{Y}_M, \mathbf{y}_M).
\end{equation}

Such a loss only considers the overall mixed image information. However, the introduced image mixing method (Sec.\,\ref{sec:max-min_attention_region_mixing}) endows objects of $\mathbf{y_A}$ and $\mathbf{y_B}$ within the content of the mixed image $\mathbf{x}_M$.
%
%With the aforementioned image mixing method in Sec.\,\ref{sec:max-min_attention_region_mixing}, the mixed image contains objects corresponding $\mathbf{y_A}$ and $\mathbf{y_B}$ in the determined regions. 
Therefore, it is plausible to supervise different regions in mixed images with different labels.  To achieve this purpose, we reshape the final output image token sequence $\mathbf{X}_M$ into image shape of
%$\mathbf{X}_{M,P} \in
$\mathbf{X}_M \in \mathbb{R}^{\frac{H}{P} \times \frac{W}{P} \times d}$ where $d$ is the final token embedding size.
Accordingly, we aggregate the tokens from the source image as:
\begin{gather}
\bar{\mathbf{X}}_{A} = \frac{1}{hw}\sum_{p,q}\mathbf{X}_{M}^{i_t+p-\lfloor \frac{h}{2} \rfloor, j_t+q-\lfloor \frac{w}{2} \rfloor},
\end{gather}
and aggregate the tokens from the target image as:
\begin{gather}
\bar{\mathbf{X}}_{B} = \frac{1}{HW-hw}\sum_{i',j'}\mathbf{X}_{M}^{i',j'},
\end{gather}
where  $(i',j') \in \big\{(i',j')|1\le i' \le \frac{H}{P}, 1\le j' \le \frac{W}{P}, (i',j') \notin \{(i_t+p-\lfloor \frac{h}{2} \rfloor, j_t+q-\lfloor \frac{w}{2} \rfloor)\} \big\}$. Then, their prediction distributions are derived from the classifier $\mathcal{F}$:
\begin{gather}
\bar{\mathbf{Y}}_A = \mathcal{F}(\bar{\mathbf{X}}_A), \nonumber \\
\bar{\mathbf{Y}}_B = \mathcal{F}(\bar{\mathbf{X}}_B).
\end{gather}

Then, SMMix supervises the fine-grained prediction distributions with fine-grained labels, $\mathbf{y}_A$ and $\mathbf{y}_B$, as:
\begin{equation}\label{eq:fine_loss}
    L_{fine} = \frac{1}{2}\big(CE(\bar{\mathbf{Y}}_A,\mathbf{y}_A) + CE(\bar{\mathbf{Y}}_B, \mathbf{y}_B)\big).
\end{equation}

Such fine-grained supervision can help ViTs locate target objects and improve their recognition ability. Besides, the additional computational costs are negligible, relying only upon the existing outputs and labels.

\subsection{Feature Consistency Constraint}\label{sec:feature_consistency_constraint}
The semantic content of the mixed image, $\mathbf{x}_M$, is equivalent to the mixing of the semantic content of the unmixed images, $\mathbf{x}_A$ and $\mathbf{x}_B$. However, the semantic content of the mixed image is more complex, increasing the difficulty of extraction of features. To help features of the mixed images fall into a consistent space with those of the original unmixed images, similar to label combination, we linearly combine the prediction distributions $\mathbf{Y}_A$ and $\mathbf{Y}_B$ of unmixed images $\mathbf{x}_A$ and $\mathbf{x}_B$, and supervise $\mathbf{Y}_M$ with the combined prediction distribution. Then, we have:
% KL loss
\begin{equation}\label{eq:con}
    L_{con} = KL\big(\mathbf{Y}_M,  \lambda_M\mathbf{Y}_A + (1-\lambda_M)\mathbf{Y}_B\big),
\end{equation}
where $KL(\cdot,\cdot)$ represents the Kullback-Leibler divergence. Note that, the prediction distributions, ${\mathbf{Y}}_A$ and ${\mathbf{Y}}_B$ in Eq.\,(\ref{eq:con}), and image attention score, $\boldsymbol{\alpha}_A$ and $\boldsymbol{\alpha}_B$ in Eq.\,(\ref{eq:center_index}), of unmixed images are obtained in the same forward propagation during training. Therefore, SMMix does not rely on pre-trained models and requires only one additional forward propagation in the training process.

\subsection{Training Objective}
Above all, in addition to the common classification loss of Eq.\,(\ref{classification}), we also require fine-grained label assignment and feature consistency constraint losses, respective in compliance with Eq.\,(\ref{eq:fine_loss}) and Eq.\,(\ref{eq:con}). Consequently, the overall training loss of our SMMix is then written as follows:
\begin{equation}
    L_{total} = L_{cls} +  L_{fine} + L_{con}.
\end{equation}

\section{Experiments}

We evaluate SMMix in four aspects: 1) Sec.\,\ref{sec:imagenet_results}, evaluating image classification task on various ViT-based architectures, 2) Sec.\,\ref{sec:downstream_tasks}, transferring pre-trained models to downstream semantic segmentation and object detection tasks, 3) Sec.\,\ref{sec:rebustness}, transferring pre-trained models to out-of-distribution datasets, 4) Sec.\,\ref{sec:performance_analysis}, exploring the quality of mixed images. Note that in the tables, our SMMix is highlighted in \hl{gray}, and \textbf{bold} denotes the best results. 

\subsection{ImageNet Classification}\label{sec:imagenet_results}

\textbf{Settings}.
We evaluate the ability of our SMMix to improve classification performance on ImageNet-1k dataset~\cite{deng2009imagenet}, which is a 1,000-class dataset, consisting of 1.28M training images and 50k validation images. 
Experiments are conducted on several recent ViT-based architectures, including DeiT~\cite{touvron2021training}, PVT~\cite{wang2021pyramid}, CaiT~\cite{touvron2021going}, and Swin~\cite{liu2021swin}. 
All models are trained on the training set, and we report the top-1 accuracy on the validation set.
For a fair comparison, we follow the implementations of the official papers.
%except for Random Erasing~\cite{zhong2020random}, which we experimentally found harmful to the proposed SMMix. 
We train all models for 300 epochs.
Both RandAugment~\cite{cubuk2020randaugment} and Mixup~\cite{zhang2017mixup} are used as default. We simply replace the original CutMix~\cite{yun2019cutmix} with the proposed SMMix, and switch SMMix and Mixup with a probability of 0.5. The image attention scores $\boldsymbol{\alpha}$ in Eq.\,(\ref{eq:image_attention_score}) are obtained from the last transformer block by feeding the unmixed images into the model under training.

\begin{table*}[!t]
\begin{minipage}{0.6\linewidth}
\begin{center}
\setlength\tabcolsep{1pt}
\begin{tabular}{ccccc>{\columncolor[gray]{0.9}}c}
\hline
% \noalign{\smallskip}
\rowcolor{white}
\multirow{2}{*}{Model} & \multirow{2}{*}{FLOPs(G)} & \multicolumn{4}{c}{Top-1 Acc.(\%)}  \\
& & CutMix & TransMix & TokenMix & SMMix \\
\hline
DeiT-T~\cite{touvron2021training} & 1.3 & 72.2 & 72.6 & 73.2 & \textbf{73.6}{\color{blue}(+1.4)} \\
DeiT-S~\cite{touvron2021training} & 4.7 & 79.8 & 80.7 & 80.8 & \textbf{81.1}{\color{blue}(+1.3)} \\
DeiT-B~\cite{touvron2021training} & 17.6 & 81.8 & 82.4 & \textbf{82.9} & \textbf{82.9}{\color{blue}(+1.1)} \\
CaiT-XXS-24~\cite{touvron2021going} & 2.5 & 77.6 & - & 78.0 & \textbf{78.9}{\color{blue}(+1.3)} \\

CaiT-XXS-36~\cite{touvron2021going} & 3.8 & 79.1 & 79.8 & - & \textbf{80.2}{\color{blue}(+1.1)} \\
PVT-T~\cite{wang2021pyramid} & 1.9 & 75.1 & 75.5 & 75.6 & \textbf{76.4}{\color{blue}(+1.3)} \\
PVT-S ~\cite{wang2021pyramid} & 3.8 & 79.8 & 80.5 & - & \textbf{81.0}{\color{blue}(+1.2)} \\
PVT-M ~\cite{wang2021pyramid} & 6.7 & 81.2 & 82.1 & - & \textbf{82.2}{\color{blue}(+1.0)} \\
PVT-L ~\cite{wang2021pyramid} & 9.8 & 81.7 & 82.4 & - & \textbf{82.7}{\color{blue}(+1.0)} \\
Swin-T ~\cite{liu2021swin} & 4.5 & 81.2 & - & 81.6 & \textbf{81.8}{\color{blue}(+0.6)} \\
\hline
\end{tabular}
\caption{Comparison of SMMix with CutMix variants designed for ViT on ImageNet-1k. ``-'' indicates that the corresponding results do not report in the original paper. Blue indicates the performance improvement compared with CutMix. }
\label{tab:baseline_compare}
\end{center}
\end{minipage}
\vspace{-1.0em}
\hspace{.1in}
\begin{minipage}{0.36\linewidth}
    
    \begin{center}
    \setlength\tabcolsep{1pt}
    \begin{tabular}{ccccc}
    \hline
    Methods         & DeiT-S~\cite{touvron2021training} & Swin-T~\cite{liu2021swin} \\ \hline
    Vanilla~\cite{li2022openmixup} & 75.7 & 80.2 \\
    CutMix~\cite{yun2019cutmix}          & 79.8   & 81.2   \\
    AttentiveMix~\cite{walawalkar2020attentive}    & 80.3   & 81.3   \\
    SaliencyMix~\cite{uddin2020saliencymix}     & 79.9   & 81.4   \\
    PuzzleMix~\cite{kim2020puzzle}       & 80.5   & 81.5     \\
    F-Mix~\cite{harris2020fmix}           & 77.4  & 79.6    \\
    % ManifoldMix~\cite{verma2019manifold} & 78.0 & 81.2 \\
    ResizeMix~\cite{qin2020resizemix}       & 78.6  & 81.4    \\
    % TokenLabel      & 80.7   &        &        \\
    AutoMix~\cite{liu2021unveiling}         & 80.8   & \textbf{81.8}     \\
    \rowcolor[gray]{0.9} SMMix (Ours)    & \textbf{81.1}   & \textbf{81.8}     &    \\ 
    \hline
    \end{tabular}
    \caption{Comparison of SMMix with other CutMix variants on ImageNet-1k. We get the performance of previous CutMix variants on DeiT-S and Swin-T from the OpenMixup~\cite{li2022openmixup} benchmark. }
    \label{tab:mix_method_compare}
    \end{center}
\end{minipage}
\end{table*}

\textbf{Results}. We first compare SMMix with recent ViT-special CutMix variants, including TransMix~\cite{chen2022transmix} and TokenMix~\cite{liu2022tokenmix}. As shown in Table\,\ref{tab:baseline_compare}, SMMix
consistently surpasses TransMix (+0.1\% $ \sim $ +1.0\%) and TokenMix (+0.2\% $ \sim $ +0.9\%) in various ViT-based architectures. In particular, SMMix can boost the top-1 accuracy by more than +1\% in DeiT-T/S/B~\cite{touvron2021training}, CaiT-XXS-24/36~\cite{touvron2021going}, and PVT-T/S/M/L~\cite{wang2021pyramid} compared with the CutMix~\cite{yun2019cutmix} baseline. Recent TokenMix~\cite{liu2022tokenmix} also achieves 82.9\% Top-1 accuracy with DeiT-B, but it re-quires a pre-trained NFNet-F6 model with 438M parameters. For models with stronger inductive bias, such as Swin-T, SMMix also provides +0.6\% performance improvement.

In Table\,\ref{tab:mix_method_compare}, we further compare SMMix with other CutMix variants, including AttentiveMix~\cite{walawalkar2020attentive}, SaliencyMix~\cite{uddin2020saliencymix}, PuzzleMix~\cite{kim2020puzzle}, F-Mix~\cite{harris2020fmix},  ResizeMix~\cite{qin2020resizemix}, and AutoMix~\cite{liu2021unveiling}. Observably, SMMix has a performance advantage over other methods. Note that SMMix is also less overhead than previous methods that require pre-trained models~\cite{walawalkar2020attentive,jiang2021all}, double forward and backward propagations~\cite{kim2020puzzle}, or additional generators~\cite{liu2021unveiling}. Specially, AutoMix~\cite{liu2021unveiling} has the same performance as our SMMix in Swin-T. However, AutoMix requires more training time (See Figure\,\ref{fig:efficiency} for detail) since AutoMix requires training an additional generator.

\begin{table}[!t]
    \centering
    \begin{tabular}{ccc}
    \hline
    Backbone & mIoU(\%) & mAcc(\%)   \\
    \hline
     PVT-T~\cite{wang2021pyramid}  & 36.6 & 46.7 \\
     \rowcolor[gray]{0.9} SMMix-PVT-T  & \textbf{37.3}{\color{blue}(+0.7)} &  \textbf{48.1}{\color{blue}(+1.4)} \\
    \hline
    PVT-S~\cite{wang2021pyramid} &  41.9 & 53.0 \\
    \rowcolor[gray]{0.9} SMMix-PVT-S & \textbf{43.0}{\color{blue}(+1.1)} & \textbf{54.1}{\color{blue}(+1.1)} \\
    \hline
    \end{tabular}
    \caption{Transferring the pre-trained models to downstream semantic segmentation task using Semantic FPN with PVT backbone on ADE20K dataset.}
    \label{tab:semantic_segmentation}
\end{table}

\begin{table}[!t]
    \setlength\tabcolsep{4pt}
    \centering
    \begin{tabular}{cccc}
    \hline
    % \noalign{\smallskip}
    Backbone & AP$^b$ & AP$^b_{50}$ & AP$^b_{75}$   \\
    \hline
    PVT-T~\cite{wang2021pyramid}  & 36.7 & 59.2 & 39.3 \\
    \rowcolor[gray]{0.9} SMMix-PVT-T  &  \textbf{37.1}{\color{blue}(+0.4)} &  \textbf{59.8}{\color{blue}(+0.6)} &  \textbf{39.6}{\color{blue}(+0.3)}\\
    \hline
   PVT-S~\cite{wang2021pyramid}  & 40.4 & 62.9 & 43.8 \\
   \rowcolor[gray]{0.9} SMMix-PVT-S &  \textbf{41.0}{\color{blue}(+0.6)} &  \textbf{63.9}{\color{blue}(+1.0)} &  \textbf{44.4}{\color{blue}(+0.6)} \\
     \hline
    \end{tabular}
    \caption{Transferring the pre-trained models to downstream object detection task using Mask R-CNN with PVT backbone on COCO val2017 dataset.}
    \label{tab:object_detection}
\end{table}

\subsection{Downstream Tasks}\label{sec:downstream_tasks}
%Pre-training on ImageNet and then fine-tuning on downstream tasks is a common practice for many visual tasks. 
To verify the generalization of our method, we also evaluate our SMMix pre-trained models on downstream tasks, including semantic segmentation and object detection. PVT~\cite{wang2021pyramid} is selected as the backbone, and we follow all training settings on PVT~\cite{wang2021pyramid} for fair comparisons.

\textbf{Semantic segmentation}.  We use ADE20K~\cite{zhou2017scene} to evaluate the performance of semantic segmentation task. ADE20k is a challenging scene parsing dataset covering 150 semantic categories, with 20k, 2k, and 3k images for training, validation, and testing. We evaluate PVT backbones with Semantic FPN~\cite{kirillov2019panoptic}. As shown in Table~\ref{tab:semantic_segmentation}, SMMix improves PVT-T for +0.7\% mIoU and PVT-S for +1.1\% mIoU.

\textbf{Object detection}. We choose the challenging COCO benchmark~\cite{lin2014microsoft} for the object detection task. All models are trained on COCO train2017 (118k images) and evaluated on val2017 (5k images). We evaluate PVT backbones with Mask R-CNN~\cite{he2017mask}. Table~\ref{tab:object_detection} shows that SMMix improves PVT-T for +0.4\% box AP, and PVT-S for +0.6\% box AP. 

These results demonstrate that the models pre-trained with the proposed SMMix consistently improve the performance on downstream tasks. Therefore, SMMix can be widely used for model training because of its excellent generalization. Note that not all augmentation-based pre-training methods bring benefits to downstream tasks. For example, CutMix~\cite{chu2021twins} has observed that pre-training with Mixup~\cite{zhang2017mixup} and CutOut~\cite{devries2017improved} failed to improve the object detection performance over the vanilla pre-trained models.

\begin{figure*}[!t]
  \centering
  \begin{subfigure}{0.3\linewidth}
    \includegraphics[height=0.95\linewidth]{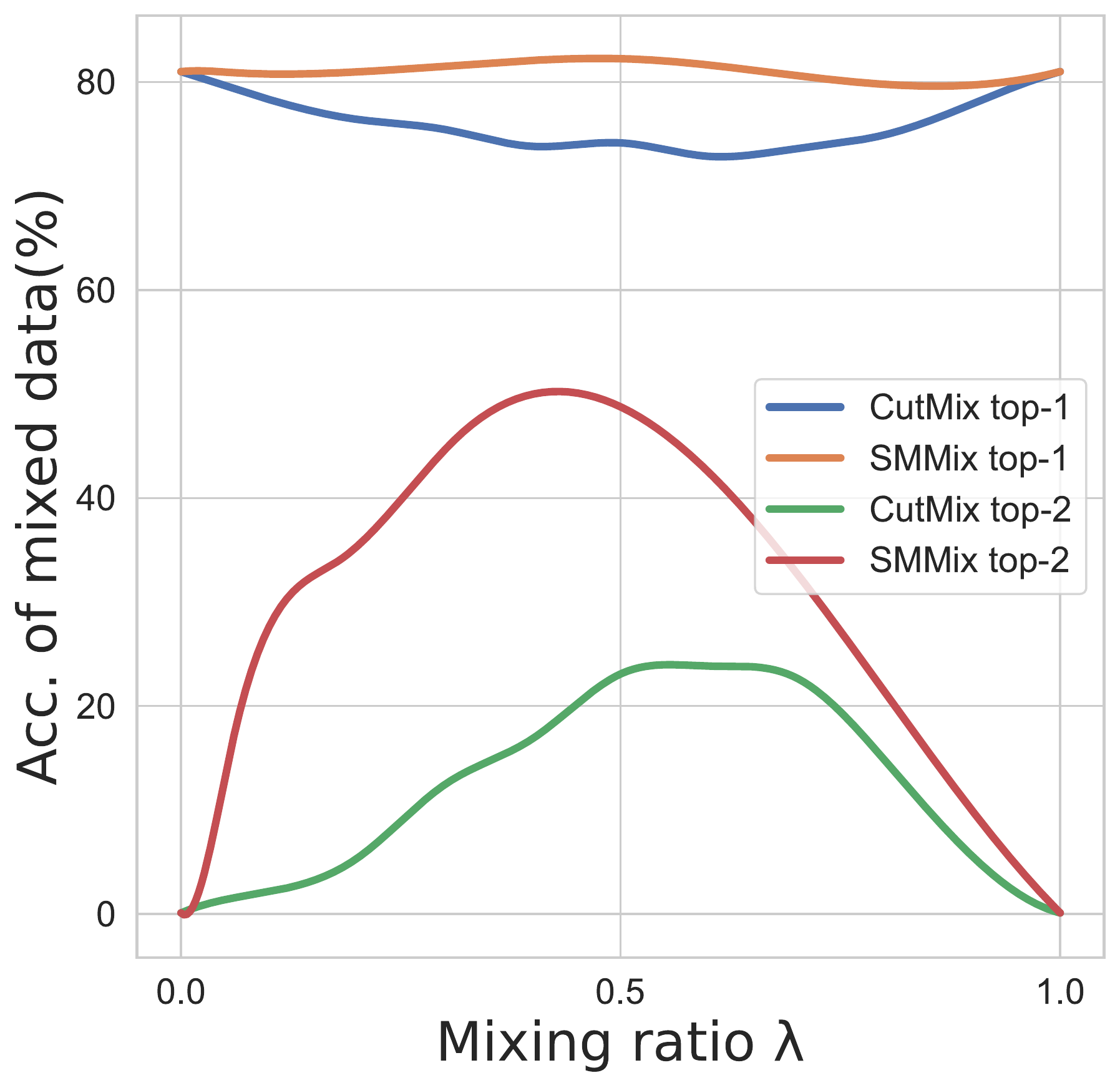}
    \caption{}
    \label{fig:top12}
  \end{subfigure}
  \hspace{0.13em}
  \begin{subfigure}{0.3\linewidth}
    \includegraphics[height=\linewidth]{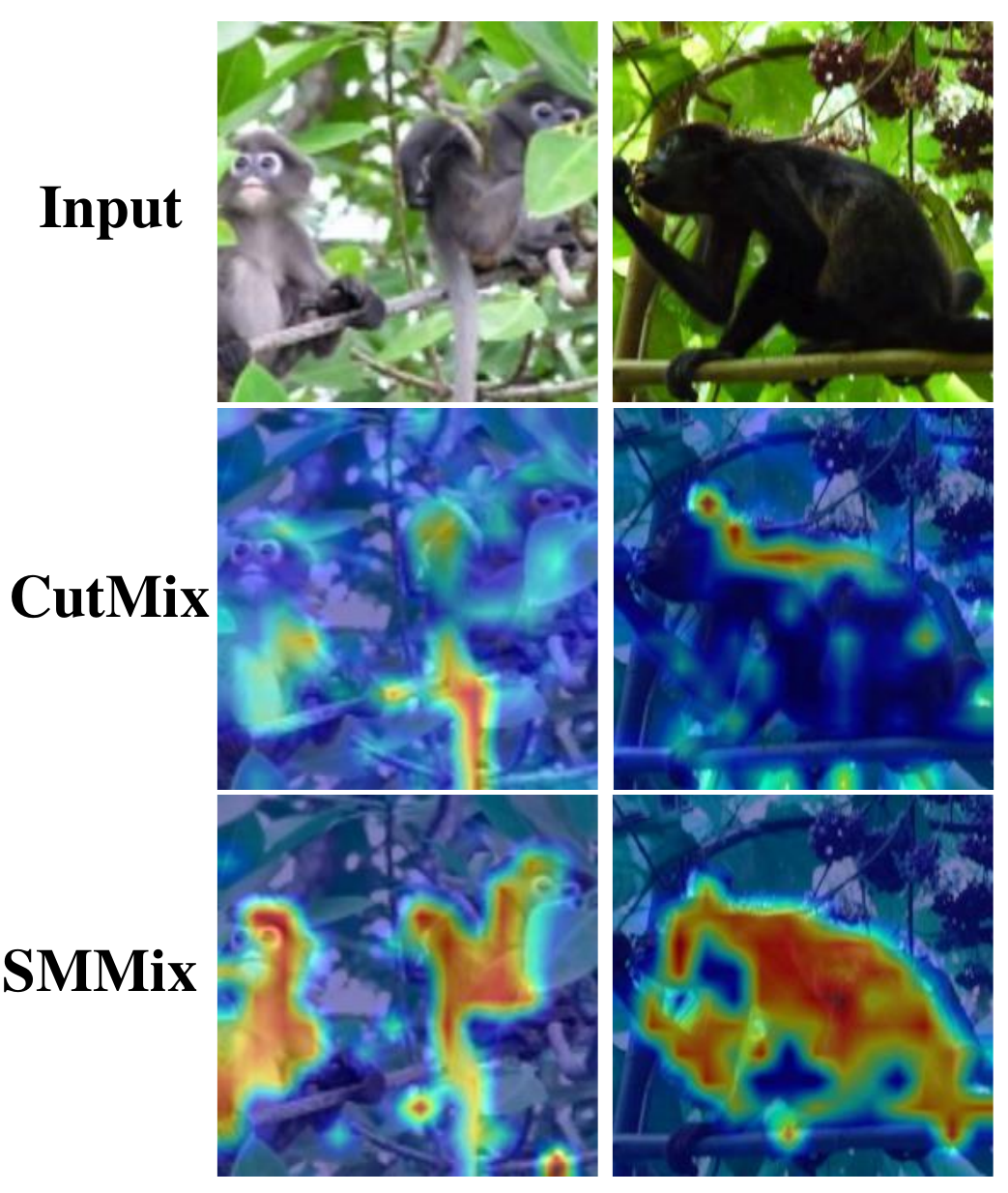}
    \caption{}
    \label{fig:visual_origin}
  \end{subfigure}
  \begin{subfigure}{0.3\linewidth}
    \includegraphics[height=\linewidth]{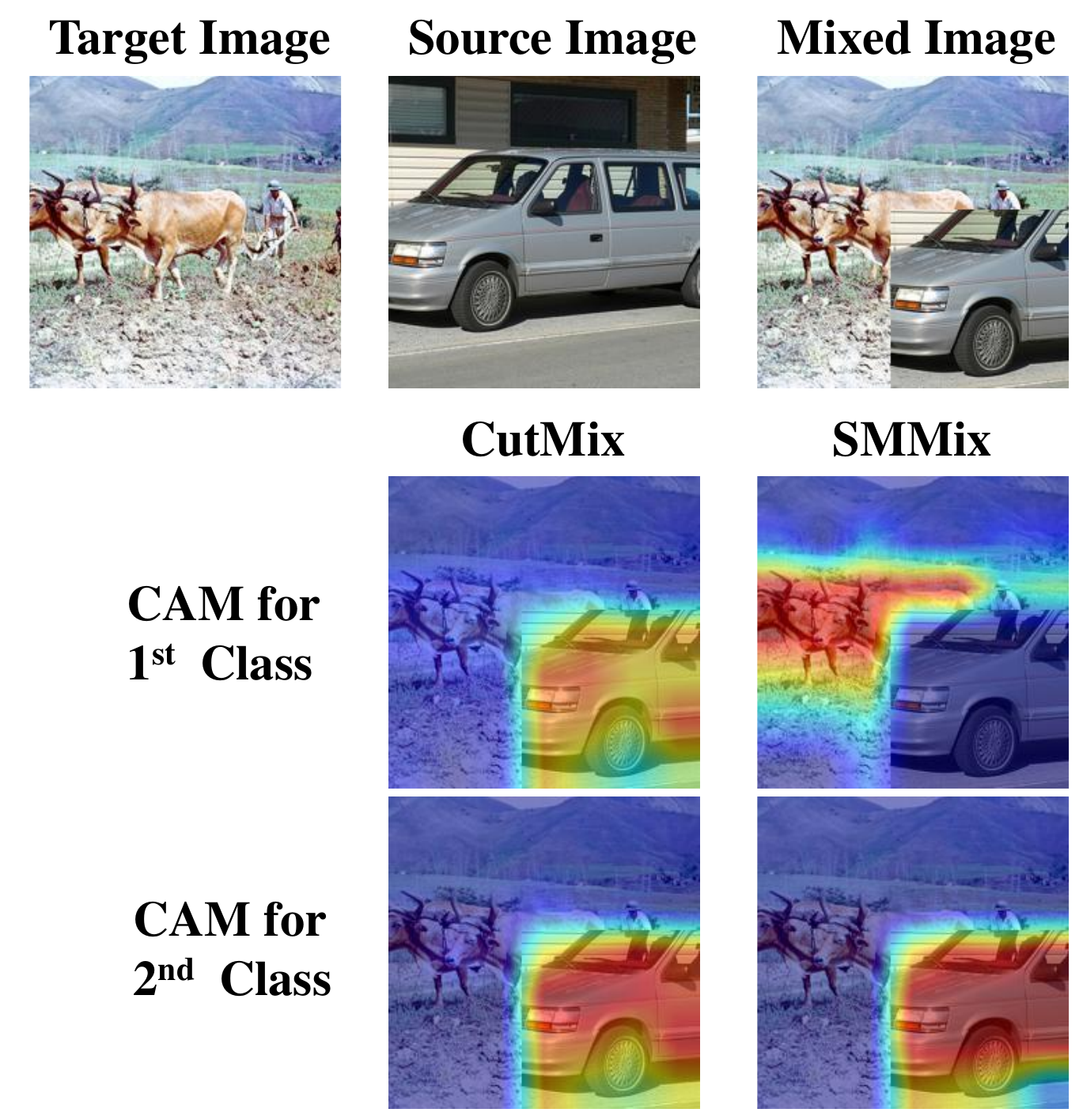}
    \caption{}
    \label{fig:visual_mixed}
  \end{subfigure}
  \vspace{-1em}
  \caption{(a) Top-1/2 accuracy of mixed images on ImageNet-1k. Top-1 accuracy is calculated by counting the top-1 prediction belongs to $\{y_A,y_B\}$. Top-2 accuracy is calculated by counting the top-2 prediction equal to  $\{y_A,y_B\}$.  (b) and (c) show the class activation map~\cite{selvaraju2017grad} of the models trained with CutMix and SMMix by testing on unmixed and mixed images, respectively.}
  \label{ablation:hyperparameters}
  \vspace{-1em}
\end{figure*}

\begin{table}[!t]
    \centering
    \setlength\tabcolsep{4pt}
    \begin{tabular}{ccccc}
    \hline
    \multirow{2}{*}{Model} & \multicolumn{4}{c}{CutMix/SMMix Top-1 Acc.(\%)} \\
     & ImageNet-A & Rendition &  Sketch & Stylized \\
     \hline
    DeiT-T & 7.1/\textbf{8.5} & 33.2/\textbf{34.9} & 20.3/\textbf{22.2} & 10.7/\textbf{11.2} \\
    DeiT-S &  18.7/\textbf{22.0} & 42.5/\textbf{43.9} & 29.5/\textbf{31.2} & 15.2/\textbf{16.6}\\
    DeiT-B &  25.2/\textbf{28.1} & 50.2/\textbf{51.7} & 36.3/\textbf{38.1} & 21.5/\textbf{22.3}\\
    % DeiT-B & /\textbf{} & /\textbf{} & /\textbf{} & /\textbf{}\\
    PVT-T & 7.7/\textbf{9.4} & 34.1/\textbf{35.2} & 21.3/\textbf{22.2} & 11.7/\textbf{12.5}  \\
    PVT-S & 17.7/\textbf{20.4} & 40.5/\textbf{41.8} & 27.1/\textbf{29.2} & 13.8/\textbf{15.4}  \\
    PVT-M & 24.8/\textbf{28.3} & 42.1/\textbf{44.4} & 30.1/\textbf{31.4} & 13.3/\textbf{15.6}  \\
    PVT-L & 26.3/\textbf{30.0} & 44.1/\textbf{44.9} & 29.9/\textbf{31.5} & 14.0/\textbf{16.3}  \\
    Swin-T & 20.7/\textbf{22.3} & 41.8/\textbf{43.1} & 29.2/\textbf{29.5} & 13.5/\textbf{13.8} \\
    \hline
    \end{tabular}
    \caption{Performance of various ViT architectures trained with CutMix/SMMix on ImageNet-1k and evaluated on four out-of-distribution datasets. Acc1/\textbf{Acc2} refers to the top-1 accuracy of the models trained with CutMix and SMMix, respectively.}
    \vspace{-1em}
    \label{tab:out_of_distribution}
\end{table}

\begin{table}[!t]
    \centering
    \begin{tabular}{ccc}
    \hline
    Model & Label Reconstruction  &  Top-1 Acc.(\%) \\
    \hline
    \multirow{3}{*}{DeiT-S~\cite{touvron2021training}} & TransMix~\cite{chen2022transmix} & 81.1 \\
    & TokenMix~\cite{liu2022tokenmix} & \textbf{81.2} \\
    & w/o (ours) & 81.1 \\
    \hline
    \end{tabular}
    \caption{The performance of DeiT-S when introducing label reconstruction methods into SMMix.}
    \label{tab:label_tuning}
    \vspace{-1em}
\end{table}

\subsection{Robustness}\label{sec:rebustness}
To verify whether SMMix can improve the robustness of ViT-based models, we also evaluate our SMMix on four out-of-distribution datasets: 
$(i)$ ImageNet-A~\cite{hendrycks2021natural} contains 7,500 adversarial examples for 200 ImageNet classes, which would yield low-confidence predictions with ResNet-50~\cite{he2016deep}.
$(ii)$ ImageNet-Rendition~\cite{hendrycks2021many} contains 30,000 image renditions (\emph{e.g.} paintings, sculpture) for 200 ImageNet classes.
$(iii)$ ImageNet-Sketch~\cite{wang2019learning} consists of sketch-like images that match the ImageNet-1k validation set in terms of category and scale.
$(iv)$ ImageNet-Stylized~\cite{geirhos2018imagenet} is created by applying AdaIN~\cite{huang2017arbitrary} style transfer to ImageNet images.
We train all models on ImageNet-1k~\cite{deng2009imagenet} training set and test them on the above out-of-distribution datasets. Table\,\ref{tab:out_of_distribution} shows that the proposed SMMix can have consistent performance gains over CutMix on the out-of-distribution data. Such results demonstrate that SMMix can enhance the robustness of the ViT-based models.

%\subsection{Property Analysis}
\subsection{Performance Analysis}\label{sec:performance_analysis}
\textbf{Premium Mixed Images.} 
The image-label inconsistency issue hinders further performance improvement of CutMix. To solve this problem, our SMMix proposes max-min attention region mixing technique, which maximizes the attentive objects in mixed images. Following AutoMix~\cite{liu2021unveiling}, we statistic the top-1/2 accuracy to verify the quality of mixed images. As shown in Figure\,\ref{fig:top12}, our SMMix significantly improves the top-1/2 accuracy of mixed images compared with CutMix. Especially for the top-2 accuracy, our SMMix achieves 48.3\% while CutMix only reaches 23.8\%. Such a substantial performance improvement demonstrates that SMMix can enrich discriminative features in mixed images. To further verify the quality of mixed images generated by SMMix, we also introduce the recent label-driven reconstruction techniques~\cite{chen2022transmix,liu2022tokenmix} into SMMix. Table\,\ref{tab:label_tuning} shows that the label reconstruction methods bring negligible performance improvement, +0\% for TransMix~\cite{chen2022transmix} and +0.1\% for TokenMix~\cite{liu2022tokenmix}. These results demonstrate that the max-min attention region mixing technique successfully alleviates the image-label inconsistency problem by maximizing the information of mixed images.
In general, SMMix generates better-quality training samples to help further improve performance.

\begin{table*}[!ht]
    \centering
    % \begin{tabular}{cccc{{\columncolor{gray}{0.9}c}c}
    \begin{tabular}{cccccc>{\columncolor[gray]{0.9}}c}
    % Module & \multicolumn{3}{c}{usage} \\
    \hline
    % \hline
    Max-Min Attention Image Mixing & \XSolidBrush  & \Checkmark & \XSolidBrush & \Checkmark  & \Checkmark  & \Checkmark \\
    Fine-grained Label Assignment & \XSolidBrush  & \XSolidBrush & \XSolidBrush & \Checkmark & \XSolidBrush &  \Checkmark\\
    Feature Consistency Constraint & \XSolidBrush  & \XSolidBrush & \Checkmark &   \XSolidBrush & \Checkmark  &  \Checkmark\\
    % \hline
    % Training Time (h) & 45.1 & 49.1 & 48.9 & 49.8 & 49.6 & 49.8 \\
    % \hline
    % Training Time & 1$\times$ & 1.09$\times$ & 1.09$\times$ & 1.10$\times$ & 1.09$\times$ & 1.10$\times$ \\
    \hline
    Top-1 Acc.(\%) & 79.8 & 80.4 & 80.3 & 80.9 & 80.8  & \textbf{81.1} \\
    \hline
    \end{tabular}
     \caption{Ablation of each component of SMMix on ImageNet-1k with DeiT-S. }
    \label{tab:module_ablation}
    \vspace{-1em}
\end{table*}

\textbf{Visualization.}
In Figure\,\ref{fig:visual_origin} and Figure\,\ref{fig:visual_mixed}, we visualize the class activation map~\cite{selvaraju2017grad} of the models trained with CutMix and SMMix. Note that we choose images that can be correctly classified by both the CutMix and SMMix models. It can be seen in Figure\,\ref{fig:visual_origin} that the SMMix model can locate objects more accurately than the CutMix model in the unmixed images. Furthermore, Figure\,\ref{fig:visual_mixed} shows that for the mixed images, the SMMix model can accurately locate  objects from two different images. On the contrary, the CutMix model focuses only on the cropped regions. The misplacement of CutMix is due to the fact that the cropped regions with sharp rectangle boundaries enhance first/second-order feature statistics, resulting in self-attention operation generating basic attention scores for cropped regions regardless of content~\cite{chen2022transmix}. However, our SMMix provides fine-grained supervision for tokens from different regions, which can help the model locate the correct region. 
In the supplementary material, we also present statistical results of image attention scores that quantitatively demonstrate the phenomena observed by visualization.

\subsection{Ablation studies}
In this section, we conduct various ablation studies to better understand SMMix. We use DeiT as the backbone, with the same training settings as described in Sec.\,\ref{sec:imagenet_results} unless otherwise specified.

\textbf{Necessity of each design}.
We first analyze the efficacy of each design in our SMMix.
Note that the fine-grained label assignment must be used in conjunction with the max-min attention region mixing. In Table\,\ref{tab:module_ablation}, we increasingly add each component to the vanilla DeiT-S training recipe, where \Checkmark and \XSolidBrush denote whether or not the corresponding component is enabled. Observably, each designed component can improve the final performance. Hence, the three designs are critical to the final performance of our SMMix.
%
%SMMix requires an additional forward propagation in the training process, which results in about 1.3$\times$ MACs compared with CutMix~\cite{yun2019cutmix}. However, Table\,\ref{tab:module_ablation} shows that SMMix leads to only 1.10$\times$ training time compared with CutMix. The actual training time is less than the theoretical MACs because training time contains the I/O times, but SMMix relies on the model under training itself and introduces fewer additional I/O. In addition, the 1.10$\times$ training time of SMMix is faster than existing methods, such as TokenMix\,(1.57$\times$)~\cite{liu2022tokenmix}, TokenLabel\,(1.59$\times$)~\cite{jiang2021all}, AutoMix\,(1.90$\times$)~\cite{liu2021unveiling}, and PuzzleMix\,(2.04$\times$)~\cite{kim2020puzzle}.

\begin{table}[!t]
    \centering
    \begin{tabular}{ccc}
    \hline
    Model & $\delta$  &  Top-1 Acc.(\%) \\
    \hline
    \multirow{3}{*}{DeiT-S~\cite{touvron2021training}} & 0.5 & 81.0 \\
    & $U$(0,1) & 81.0 \\
    & $U$(0.25,0.75) & \textbf{81.1} \\
    \hline
    \end{tabular}
    \caption{Ablation of side ratio $\delta$. $U$ means uniform distribution.}
    \label{tab:side_ratio}
    \vspace{-1.5em}
\end{table}

% \begin{table}[!t]
%     \centering
%     \begin{tabular}{ccc}
%     \hline
%     Method & Epoch  &  Top-1 Acc.(\%) \\
%     \hline
%     \multirow{2}{*}{CutMix~\cite{yun2019cutmix}} & 300 & 79.8 \\
%      & 400 & 80.0{\color{blue}(+0.2)} \\
%      \hline
%     % \multirow{2}{*}{TokenMix~\cite{liu2022tokenmix}} & 300 & 80.8 \\
%     %  & 400 & 81.2{\color{blue}(+0.4)} \\
%     %  \hline
%      \multirow{2}{*}{SMMix (ours)} & 300 & 81.1 \\
%      & 400 & 81.5{\color{blue}(+0.4)} \\
%     \hline
%     \end{tabular}
%     \caption{Ablation of training epochs. }
%     \label{tab:long_epoch}
% \end{table}

\textbf{Side ratio $\delta$ of cropped rectangle}. 
$\delta$ determines the size of the cropped rectangle, which indicates the strength of regularization. We test three strategies: 1) fixed as 0.5, 2) sampled from $U$(0.25, 0.75), 3) sampled from $U$(0,1). Table\,\ref{tab:side_ratio} shows that three strategies achieve similar performance, which means that the proposed SMMix is robust to the side ratio $\delta$. We simply sample $\delta$ from the uniform distribution (0.25,0.75) by default.

% \textbf{Training epochs}.
% %
% Table\,\ref{tab:long_epoch} presents the results of longer training epochs. We can see that CutMix~\cite{yun2019cutmix} trained with 400 epochs only brings +0.2\% performance improvement over training with 300 epochs, still much lower than SMMix trained with 300 epochs. Furthermore, SMMix improves DeiT-S by +0.4\% with 100 additional training epochs, which benefits more from the longer training.

\textbf{Pre-trained models}.
An essential advantage of our work is that SMMix relies entirely on the model under training itself, \emph{i.e.}, no extra pre-trained models are required. 
Specially, SMMix first forwards the unmixed images to obtain the corresponding image attention score and prediction distribution for guiding the formal training process. We consider the forwarded model before formal training as a motivated model. Therefore, whether a pre-trained motivated model can provide better guidance than the model under training remains a question.  
For this purpose, we train DeiT-T with three motivated models: the model under training (self), pre-trained DeiT-T, and pre-trained DeiT-S. Table\,\ref{tab:different_motivated} shows that a larger-scale pre-trained model can further improve performance, while a pre-trained model on the same scale as the model under training does not provide any benefit.
Thus, it is noteworthy that we use the self-motivated paradigm without pre-train models for light training overhead. However, the proposed training technique can perform better with a larger pre-trained model, which demonstrates the potential of SMMix.

\begin{table}[!t]
    \centering
    \setlength\tabcolsep{3pt}
    \begin{tabular}{cccc}
    \hline
    Model & Motivated & Pre-trained &  Top-1 Acc.(\%) \\
    \hline
    \multirow{3}{*}{DeiT-T~\cite{touvron2021training}} & DeiT-T & \Checkmark & 73.5 \\
     & DeiT-S & \Checkmark & \textbf{74.1} \\
     & self (ours) & \XSolidBrush & 73.6 \\
    \hline
    \end{tabular}
    \caption{The performance of DeiT-T with different motivate models. The self means SMMix proposed in this paper.}
    \label{tab:different_motivated}
    \vspace{-1.5em}
\end{table}

\section{Conclusion}
This paper proposes SMMix, a novel and effective image mixing technique. Specially, we design a self-motivated paradigm that motivates both the image and label enhancement in image mixing by the model under training itself. Thus, SMMix is more flexible and easier to use than the existing CutMix variants because it has a light training overhead and eliminates the reliance on pre-trained models.
Extensive experiments verify the generalization and effectiveness of SMMix, which can significantly improve the performance of various ViT-based models. Besides, SMMix also exhibits transferability on downstream tasks and robustness to out-of-distribution datasets. Overall, we hope that the self-motivated paradigm introduced by SMMix can provide a new perspective on image mixing techniques and even on deep neural network training.

\textbf{Limitation}. We further discuss unexplored limitations, which will be our future focus. First, SMMix somewhat increases the training overhead compared with vanilla CutMix due to the need for extra forward propagation. Second, SMMix is based on the self-attention and patch-splitting operation of ViTs. More efforts can be made to transfer the idea of SMMix to convolutional neural networks. 

\section*{Acknowledgement}
\begin{sloppypar}
This work was supported by National Key R\&D Program of China (No.2022ZD0118202), the National Science Fund for Distinguished Young Scholars (No.62025603), the National Natural Science Foundation of China (No. U21B2037, No. U22B2051, No. 62176222, No. 62176223, No. 62176226, No. 62072386, No. 62072387, No. 62072389, No. 62002305 and No. 62272401), and the Natural Science Foundation of Fujian Province of China (No.2021J01002,  No.2022J06001).
\end{sloppypar}

{\small
\bibliographystyle{ieee_fullname}
\bibliography{egbib}
}

\appendix

\renewcommand{\thetable}{A\arabic{table}}
\renewcommand{\thefigure}{A\arabic{figure}}
\renewcommand{\theequation}{A\arabic{equation}}
\renewcommand{\thesection}{A\arabic{section}}

\setcounter{table}{0}
\setcounter{figure}{0}

\begin{table*}[!ht]
    \hsize = \textwidth
    \centering
    \renewcommand{\arraystretch}{1.5}
    \begin{tabular}{ccc}
    % \linespread{2} 
    \hline
    \diagbox{Query}{Key}   & Source & Target  \\
    \hline
        Source & $\frac{1}{N_s^2}\sum \mathbf{A}^{i,j}, s.t.\text{ } i \in \mathbf{I}_s, j \in \mathbf{I}_s  $ & 
        $\frac{1}{N_s N_t}\sum \mathbf{A}^{i,j}, s.t.\text{ } i \in \mathbf{I}_s, j \in \mathbf{I}_t$ \\
        Target & $\frac{1}{N_s N_t}\sum \mathbf{A}^{i,j}, s.t.\text{ } i \in \mathbf{I}_t, j \in \mathbf{I}_s$ & 
        $\frac{1}{N_t^2}\sum \mathbf{A}^{i,j}, s.t.\text{ } i \in \mathbf{I}_t, j \in \mathbf{I}_t$ \\
    \hline
    \end{tabular}
    \caption{Calculation detail of average attention scores between image tokens from different regions. }
    \label{tab:calculation_process}
\end{table*}

\section{Statistics of Attention Score} \label{sec:attention_ststistics}
Attention score can reflect the similarity of each token to the others. For mixed images, tokens from the same unmixed image are more similar than those from different unmixed images. To further corroborate the visualization in Figure\,\ref{fig:visual_mixed} of the main paper with quantitative data, we calculate average attention scores among image tokens from different regions of the mixed images.

\textbf{How to calculate?} We leverage SMMix to generate new mixed images based on ImageNet-1k~\cite{deng2009imagenet}. A mixed image  contains two regions respectively from the source and target images.  The mixed image is fed into a ViT to obtain a token sequence, $\mathbf{T} \in \mathbb{R}^{N \times d}$. For a simpler representation, we simply assume that $\mathbf{I}_s = \{\mathbf{I}_s^1, \mathbf{I}_s^2, ..., \mathbf{I}_s^{N_s}\}$ and $\mathbf{I}_t = \{\mathbf{I}_t^1, \mathbf{I}_t^2, ..., \mathbf{I}_t^{N_t}\}$, where $\mathbf{I}_s$ and $\mathbf{I}_t$ are the indexes of the tokens from the source and target regions, respectively;
$N_s$ and $N_t$ respectively indicate the token number of the source and target regions, and $N_s + N_t = N$.
Following Eq.\,(\ref{eq:attention}) in the main paper, we obtain the self-attention matrix, $\mathbf{A} \in \mathbb{R}^{N \times N}$, which contains attention scores among each token; $\mathbf{A}^{i,j}$ denotes the attention score when taking the $i$-th token as a query and the $j$-th token as a key. There are two types of tokens, either from the source or target region. Thus, self-attention forms four (query, key) pairs for mixed images according to the token region. Table\,\ref{tab:calculation_process} shows how to calculate average attention scores for the four (query, key) pairs.

\begin{table}[ht]
    \caption{Attention scores among image tokens from source/target regions. Intuitively, tokens should pay more attention to the tokens from the same regions.  Score1/$\mathbf{Score2}$ refers to corresponding attention scores of the models trained with CutMix and SMMix, respectively. }
    \centering
    \begin{tabular}{ccc}
    \hline
    \diagbox{Query}{Key}   & Source & Target  \\
    \hline
        Source &  0.0122/\textbf{0.0142}{\color{blue}$\uparrow$} & 0.0037/\textbf{0.0031}{\color{blue}$\downarrow$} \\
        Target & 0.0098/\textbf{0.0021}{\color{blue}$\downarrow$} & 0.0046/\textbf{0.0070}{\color{blue}$\uparrow$} \\
    \hline
    \end{tabular}
    \label{tab:region_level_attention_score}
\end{table}

\textbf{Results}. We can find two interesting phenomena in Table\,\ref{tab:region_level_attention_score}:

First, SMMix assists tokens focus more on tokens from the same regions. For example, when both the query and key tokens are from the same regions, the SMMix pre-trained model has attention scores of $0.0142$ and $0.0070$, which are higher than the CutMix pre-trained model's $0.0122$ and $0.0046$.

Second, SMMix alleviates incorrect attention scores caused by sharp rectangle boundaries. Taking tokens from target regions as queries, we find that the CutMix pre-trained model focuses more on tokens from source regions ($0.0098$) than tokens from target regions ($0.0046$). The incorrect attention scores are caused by sharp rectangle boundaries, which enhance the first/second-order feature statistics and cause self-attention operation to generate basic
attention scores for the cropped rectangles regardless of contents. However, taking tokens from target regions as a query, SMMix pre-trained models successfully focus more on tokens from target regions ($0.0070$), rather than tokens from source regions ($0.0021$).

These two phenomena show that ViTs pre-trained with SMMix can generate more appropriate attention scores and help the model locate the accurate regions.

\section{Additional Results}\label{sec:additional_results}

\textbf{Comparisons with TokenLabel.} Tabel\,\ref{tab:tokenlabel} compares our SMMix with TokenLabel~\cite{jiang2021all}. We observe that SMMix outperforms TokenLabel in DeiT-T (+0.7\%) and DeiT-S (+0.1\%). Also, SMMix has less training time and without dependence on any pre-trained models, while TokenLabel requires a NAFNet-F6 model~\cite{brock2021high} that has 438M parameters.

%Therefore, SMMix not only has better performance but also has less training overhead compared to TokenLabel.

\textbf{Variants of max-min attention region mixing}.
For the max-min attention region mixing, we select the maximum-scored region from a source image and paste it to the minimum-scored region in a target image. Such an operation can maximize the information of mixed images and make the proposed fine-grained label assignment feasible. To demonstrate the effectiveness of such a mixing pattern, we consider five possible variants:
\begin{itemize}
    \item $(i)$ Random $\to$ Corr: randomly select a region from the source image and paste it to the same location in the target image;
    \item $(ii)$ Random $\to$ Max Attn: randomly select a region from the source image and pastes it to the maximum-scored region in the target image;
    \item $(iii)$ Random $\to$ Min Attn: which randomly select a region from the source image and paste it to the minimum-scored region in the target image;
    \item $(iv)$ Max Attn $\to$ Corr: which select the maximum-scored region from the source image and paste it to the  same location in the target image;
    \item $(v)$ Max Attn $\to$ Max Attn: which select the maximum-scored region from the source image and paste it to the maximum-scored region in the target image.
\end{itemize}

Finally, we denote our max-min attention region mixing as Max Attn $\to$ Min Attn.
Table\,\ref{tab:region_selection_ablation} compares the performance. Obviously, our $Max\text{ }Attn \to Min\text{ }Attn$ achieves the best performance compared to its variants, because it maximizes the information of mixed images. On the other hand, $Random \to Max\text{ }Attn$ performs the worst, since it occludes the most targets. 
Note that these findings are inconsistent with SalinencyMix~\cite{uddin2020saliencymix}, which believes that $Attn \text{ } \to Corr$ pattern performs best since the pattern provides a trade-off between regularization and image information. We attribute the difference to two possible causes:
(1) Our image attention score locates objects more accurately than the salience detector in SaliencyMix~\cite{uddin2020saliencymix}; 
(2) The regularizations strategies in the ViTs training recipe allow more information to be retained in mixing methods.

\begin{table}[!t]
\centering
    \caption{Comaprison of our SMMix with TokenLabel on ImageNet-1k. ``Pra-trained'' indicates whether to adopt a pre-trained model for the network training. ``Time'' refers to the training time increase over CutMix. }
    \centering
    \setlength\tabcolsep{0.5pt}
    \begin{tabular}{ccccc}
    \hline
    Model  & Method & Pre-trained & Time&  Top-1 Acc.(\%) \\
    \hline
    \multirow{3}{*}{DeiT-T~\cite{touvron2021training}} & Baseline & \XSolidBrush & 1.00$\times$ & 72.2 \\
     & TokenLabel & \Checkmark & 1.59$\times$ & 72.9 \\
     &\cellcolor[gray]{0.9} SMMix (ours) &\cellcolor[gray]{0.9} \XSolidBrush & \cellcolor[gray]{0.9}1.10$\times$&\cellcolor[gray]{0.9}  \textbf{73.6} \\
     \hline
    \multirow{3}{*}{DeiT-S~\cite{touvron2021training}} & Baseline & \XSolidBrush & 1.00$\times$ & 79.8 \\
     & TokenLabel & \Checkmark & 1.59$\times$ & 81.0 \\
     &\cellcolor[gray]{0.9} SMMix (ours) &\cellcolor[gray]{0.9} \XSolidBrush &  \cellcolor[gray]{0.9} 1.10$\times$ &\cellcolor[gray]{0.9} \textbf{81.1} \\
    \hline
    \end{tabular}
    \label{tab:tokenlabel}
\end{table}

\begin{table}[!t]
\caption{Ablation of different image mixing schemes on DeiT-S. All the models are trained for 100 epochs.  }
    \centering
    \begin{tabular}{cc}
    \hline
    Mixing Scheme & Top-1 Acc.(\%) \\
    \hline
    Random $\to$ Corr & 74.2 \\
    Random $\to$ Max Attn & 73.8 \\
    Random$\to$ Min Attn & 74.5 \\
    % Non Attn $\to$ Corr & 73.6 \\
    % Non Attn $\to$ Attn \\
    % Non Attn $\to$ Non Attn 74.73 \\
    Max Attn $\to$ Corr & 74.4 \\
    Max Attn $\to$ Max Attn & 74.3 \\
    \rowcolor[gray]{0.9}
    Max Attn $\to$ Min Attn(Ours) & \textbf{74.7} \\
    \hline
    \end{tabular}
    \label{tab:region_selection_ablation}
\end{table}

\begin{table}[!t]
    \setlength\tabcolsep{2pt}
    \caption{Ablation of image attention score generation. SM-Mix uses the image attention score output by the $d$-th block of DeiT-S. ``None'' means to randomly generate the image attention score. ``Rollout'' means to average all-block image attention scores.}
    \centering
    \begin{tabular}{ccccccc}
    \hline
         d &  None & 3 & 6 & 9 & 12 & Rollout  \\
        \hline
         Top-1 Acc.(\%) & 80.7 & 81.0 & 81.1 & 81.1 & 81.1 & 81.1 \\
    \hline
    \end{tabular}
    \label{tab:attention_score}
\end{table}

\textbf{Image Attention Score}. Table\,\ref{tab:attention_score} shows the performance for image attention scores from different depths. We observe 0.4\% performance drop when taking a random image attention score, demonstrating the guidance ability of the image attention score in the image mixing process. SMMix achieves the best performance when $d=6,9,12$. We set $d=12$ as the default since the feature consistency constraint requires a complete forward propagation. However, this shows that when the feature consistency constraint is disabled, SMMix can further reduce training costs by using the shallower-layer image attention score.

\begin{figure*}[!t]
    \vspace{15pt}
    \centering
    \includegraphics[width=\linewidth]{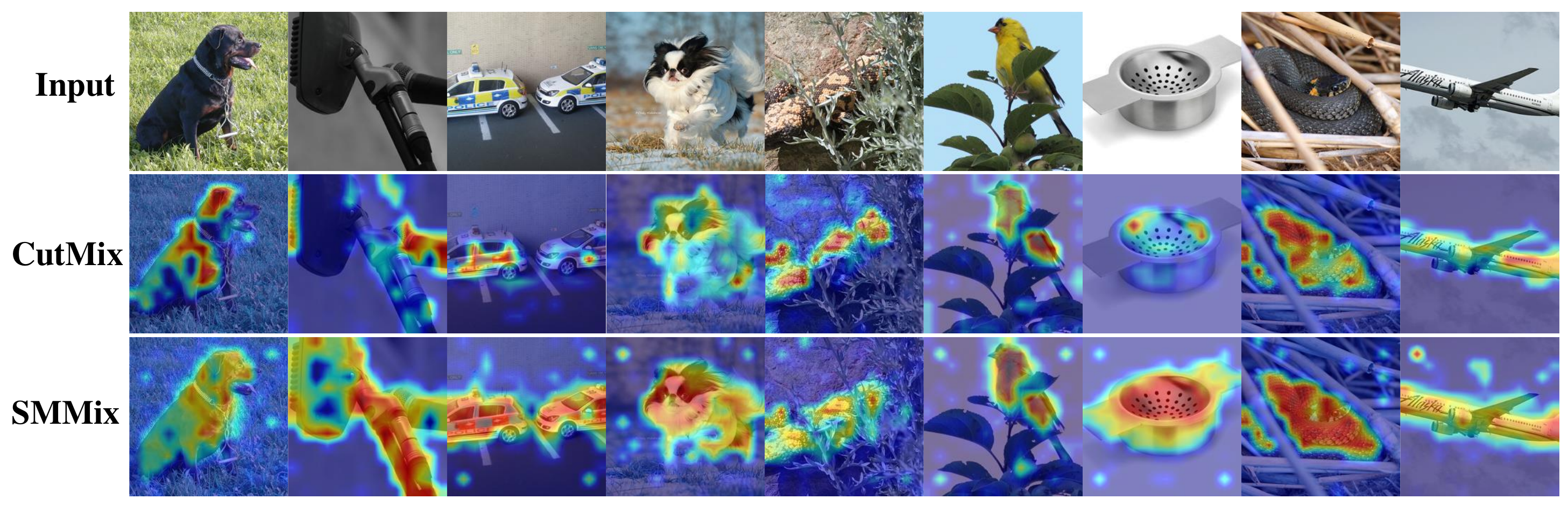}
    \caption{The class activation map~\cite{selvaraju2017grad} of the models trained with CutMix and SMMix and tested on unmixed images.}
    \label{fig:cam_single_2}
    \vspace{5pt}
\end{figure*}

\begin{figure*}[!t]
    \centering
    \includegraphics[width=\linewidth]{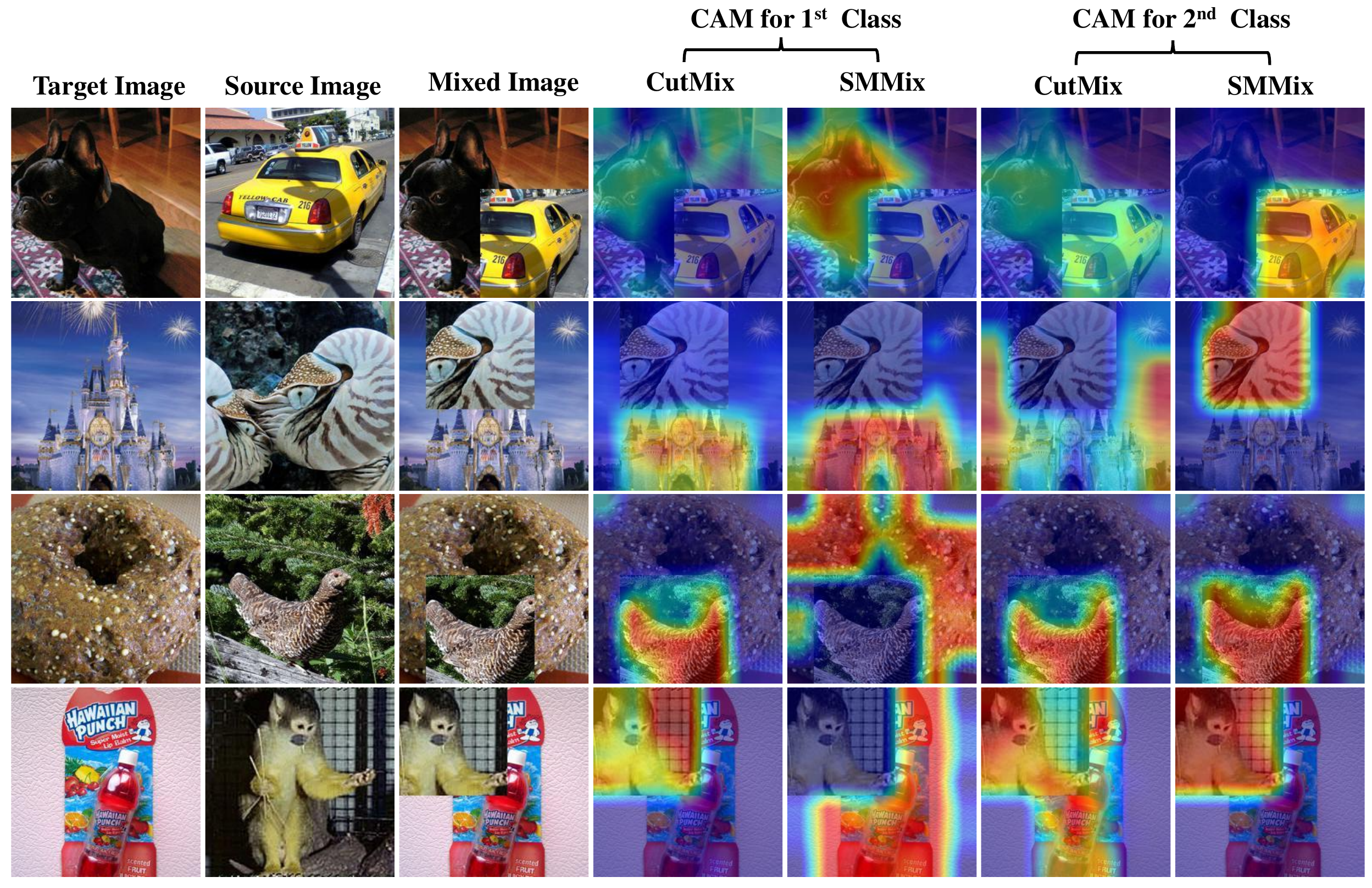}
    \caption{The class activation map~\cite{selvaraju2017grad} of the models trained with CutMix and SMMix and tested on mixed images.}
    \label{fig:cam_mixed_2}
\end{figure*}

\section{Details of Training Time Testing}\label{sec:test_detail}
In Figure\,\ref{fig:efficiency} of the main paper, we report the training time of DeiT-S~\cite{touvron2021training} on different CutMix variants. All models are trained on ImageNet-1k with a 4$\times$A100 GPU machine for 300 epochs, and AMP~\cite{micikevicius2017mixed} is activated during the training process. In particular, we follow the original DeiT training recipe except for TransMix~\cite{chen2022transmix}. Following the open source code of TransMix~\cite{chen2022transmix}, we reproduce it by modifying the batch size from 1024 to 256.

\section{More Visualization}\label{sec:more_visulization}
Figure\,\ref{fig:cam_single_2} and Figure\,\ref{fig:cam_mixed_2} provide more visual examples in ImageNet-1k. The visualization shows that models trained with our SMMix can locate objects more accurately in both unmixed and mixed images.

\end{document}